\documentclass[conference]{IEEEtran}
\IEEEoverridecommandlockouts
\usepackage{cite}
\usepackage{amsmath,amssymb,amsfonts}
\usepackage{algorithmic}
\usepackage{graphicx}
\usepackage{textcomp}
\usepackage{xcolor}
\usepackage{subfigure}
\usepackage{booktabs} % for professional tables
\usepackage{tabularx}
\usepackage{float}
\usepackage{microtype}
\usepackage{color, colortbl}
\usepackage[ruled,vlined]{algorithm2e}
\usepackage{url}
\usepackage{algorithmic}
\usepackage{mathtools}
\usepackage{dblfloatfix}
\def\BibTeX{{\rm B\kern-.05em{\sc i\kern-.025em b}\kern-.08em
		T\kern-.1667em\lower.7ex\hbox{E}\kern-.125emX}}
\begin{document}
	
	\title{Adaptive Beam Search to Enhance On-device Abstractive Summarization
	\thanks{978-1-6654-4175-9/21/\$31.00 ©2021 IEEE
}}
	
	\author{\IEEEauthorblockN{Harichandana B S S}
		\IEEEauthorblockA{\textit{Samsung R\&D Institute} \\
			Bangalore, India \\
			hari.ss@samsung.com}
		\and
		\IEEEauthorblockN{Sumit Kumar}
		\IEEEauthorblockA{\textit{Samsung R\&D Institute} \\
			Bangalore, India \\
			sumit.kr@samsung.com}
		
	}
	
	\maketitle

	\begin{abstract}
	We receive several essential updates on our smartphones in the form of SMS, documents, voice messages, etc. that get buried beneath the clutter of content. We often do not realize the key information without going through the full content. SMS notifications sometimes help by giving an idea of what the message is about, however, they merely offer a preview of the beginning content. One way to solve this is to have a single efficient model that can adapt and summarize data from varied sources. In this paper, we tackle this issue and for the first time, propose a novel Adaptive Beam Search to improve the quality of on-device abstractive summarization that can be applied to SMS, voice messages and can be extended to documents. To the best of our knowledge, this is the first on-device abstractive summarization pipeline to be proposed that can adapt to multiple data sources addressing privacy concerns of users as compared to the majority of existing summarization systems that send data to a server. We reduce the model size by 30.9\% using knowledge distillation and show that this model with a 97.6\% lesser memory footprint extracts the same or more key information as compared to BERT.
	\end{abstract}
	

	\section{Introduction}
	\label{sec:intro}
	
	Summarization is a significant challenge of natural language understanding \cite{jacquenet2019meeting}.	
	There are mainly two techniques of summarization, i.e., extractive and abstractive methods. Most successful summarization systems utilize extractive approaches that crop out and stitch together portions of the source to produce a summary \cite {nallapati2017summarunner}. In contrast, abstractive summarization systems generate new phrases, possibly rephrasing or using words that were not in the original source \cite{ chopra2016abstractive}. Extractive text summarization is easier to build however summarization by abstractive technique is superior because they produce better semantically related summary \cite{munot2014comparative}. Therefore, we choose abstractive summarization.
	
	Gartner predicts that by 2022, 80\% of smartphones shipped will have on-device AI capabilities \cite{gartner}. Though various cloud-based models are rendering domain-based summaries, most of them are not fit to be deployed on mobile devices due to their large size. Most of the current text summarization applications send the extracted text to the server to get its summarized version leading to privacy issues. On-device summarization has been explored before but most of these use extractive approach to build summaries \cite{ cabral2015automatic } \cite{ chongtay2018responsive }.
	
	On-device deployment is a significant challenge due to limited availability of memory and computational power \cite{inproceedings}. However, one of the challenges with summarization itself is that it is hard to generalize. For example, summarizing a news article is very different from summarizing a financial earnings report. Certain text features like document length or genre (tech, sports, travel, etc.) make the task of summarization a serious data science problem to solve. For this reason, we aim to build an efficient model that can adapt to many sources thus enabling a single model to work for multiple use cases and saving the memory required otherwise by deploying multiple models for the same. To tackle the generalization issue, we explore solutions on the decoding side. 
	
	\begin{table*}[!b]
		\caption{SMS Summarization results using different variants of Beam Search highlighting wrong tokens in \textcolor{red}{red} and correctly identified key points in \textcolor{blue}{blue}}
		\label{t1}
		\vskip 0.15in
		\begin{center}
			\begin{small}
				
				\begin{tabular}{m{0.27\textwidth}|m{0.15\textwidth}|m{0.15\textwidth}|m{0.15\textwidth}|m{0.15\textwidth}}
					\hline
					
					\vskip 0.025in \textbf{Source}  & \vskip 0.025in \textbf{Taditional Beam Search}   &  \vskip 0.025in \textbf{Beam Search with keyword based stepwise increment}   &  \vskip 0.025in \textbf{Beam Search with keyword based end increment} & \vskip 0.025in \textbf{Adaptive Beam Search (ABS)} 	  \\
					\hline
					\vskip 0.025in
					Dear Member, EPF Contribution of Rs. 13458 against UAN 100385505604 for due month 122016 has been received. Passbook will be updated shortly. Regards EPFO  
					\vskip -0.025in 
					& \textcolor{red}{passbook} contribution of rs. 13458 against uan 100385505604 for due  
					& \textcolor{red}{passbook} contribution of rs. 13458 against uan 100385505604 for due
					& \textcolor{red}{passbook} contribution of rs. 13458 against uan 100385505604 for d
					&  dear member , \textcolor{blue}{epf} contribution of rs. 13458 against uan
					
					\\
					\hline
					\vskip 0.025in
					Buy Gated Farmhouse near Bangalore @ 12.5 Lakhs 45 Min frm Airport on NH 7 Developed, Maintained and Affordable Excellent Appreciation WhatsApp: trkk.in/Ua/3mR 	\vskip -0.025in
					& 45 min frm airport on nh 7 .
					& \textcolor{red}{lakhs / ua/3mr} buy gated farmhouse near bangalore @ 12.5 lakhs .
					& buy gated farmhouse near bangalore @ 12.5 lakhs . developed , maintained
					& buy gated farmhouse near bangalore @ 12.5 lakhs . developed , maintained
					
					\\
					
					\hline
					\vskip 0.025in
					
					Txn of INR 209.00 done on Acct XX013 on 15-DEC-19.Info: VPS*KAI P.Avbl Bal:INR 50,698.72.Call 18002662 for dispute or SMS BLOCK 013 to 9215676766 	\vskip -0.025in
					&call 18002662 for dispute or sms block 013 to 9215676766 .  
					& txn \textcolor{red}{* kai p.avbl} : inr acct . call 18002662 for dispute 
					& txn \textcolor{red}{50,698.72} of inr 209.00 done on acct xx013 on 15-dec-19 .
					&txn of \textcolor{blue}{inr 209.00} done on acct xx013 on 15-dec-19 . info					
					\\
					\hline
				\end{tabular}
			\end{small}
		\end{center}
		\vskip -0.1in
	\end{table*}
	A popular decoding technique used by seq2seq models is Beam Search \cite{graves2012sequence}. However, we find that traditional beam search does not always give consistent and meaningful results during summarization. We introduce our novel Adaptive Beam Search in Section \ref{ABS}. A significant contribution of our proposed method is summarizing not only long but also short text.
	
	Short text (like SMS) summary is relatively unexplored as compared to long text summarization. There are multiple applications of short text summary on mobile devices and wearables to provide crisp notifications owing to their small screen size. As per a recent survey by Perficient \cite{perficient} on mobile voice usage, over 19\% of users prefer to dictate messages, and over 5\% users send voice messages, making speech summarization an important task to improve user experience.
	
	In this paper, we present an abstractive on-device summarization method that can be used for a variety of mobile applications using an Adaptive Beam Search (ABS) mechanism.  
	The main contributions of this paper are as follows:
	\begin{itemize}
		\item We propose for the first time, an on-device architecture for abstractive summarization of short text, long text, and speech using a novel adaptive Beam Search strategy.
		\item We propose a novel method of optimal summary selection using Beam Search.
		\item We reduce model size using Knowledge Distillation and evaluate its effect on model performance.
		\item We benchmark short text summarization performance of our model with BERT \cite{devlin2018bert} and also evaluate overall text and speech summarization performance on CNN-DM \cite{hermann2015teaching} and How2 \cite{sanabria2018how2} datasets respectively.
		
	\end{itemize}
	The experiments on our architecture show up to a 13\% improvement in the number of keywords extracted in the final summary compared with BERT, and with a 97.6\% decrease in model-size comparatively.

	\section{Background}
	
	Summarization has been used for various applications. \cite{rush2015neural} first introduced a neural attention seq2seq model with an attention-based encoder and a neural language model augmented decoder to generate an abstractive summary achieving significant improvement over conventional methods. The following years have extended the model developed using recurrent decoders, and variational autoencoders \cite{miao2016language}. \cite{see2017get} used pointer networks \cite{vinyals2015pointer} \cite{xu2015show}. With recent advancement in the use of pre-trained language models \cite{zhang2019hibert} for Natural Language Processing(NLP) tasks, Bidirectional Encoder Representations(BERT) \cite{devlin2018bert} was used to encode the document and generate an extractive summary by stacking several inter-sentence Transformer layers. \cite{icassp} introduces new technique for speech summarization. Past research on Beam search include some works like \cite{freitag2017beam}. There has been research done to optimize this further for the
	required use case In \cite{8851891}.  But, this increases the inference time, is not feasible to deploy on-device and does not help increase generalizability as it depends on source. Thus we propose a novel technique	to overcome these issues.

	\section{Traditional Beam Search}
	Beam Search \cite{graves2012sequence} is a search strategy that tries to choose better results from all possible candidates efficiently.
	We observe that the traditional Beam Search strategy does not always yield semantically correct summaries. We experimented using a pointer-generator network  \cite{see2017get} with the coverage and traditional Beam Search with a beam width equal to 4 (determined after experimentation to optimize inference time) on short text (SMS) data. The results are shown in Table \ref{t1}. It is seen that this often results in misleading summaries when used on data other than news  (as it is trained only on news data CNN-DM, as mentioned in Section \ref{dataset}). For example, in the third example in Table \ref{t1}, the summary does not describe the actual message and is misleading. To improve the performance we experiment using a pointer-generator with better Beam Search strategies.
		
	\section{Search Strategies experimented}
	Since we aim to create a single model that can generalize across varied types of sources, we experiment with a strategy in which source-specific keywords are given more priority by increasing the log-probabilities of all hypotheses containing these according to their category. To obtain these keywords, we created an in-house dataset consisting of source files in the form of SMS, and documents belonging to different categories (OTPs, travel, etc.) (categories were determined through an internal survey). The dataset consists of 100 files of each category collected from $>$100 individuals from diverse backgrounds. We consider the top 30 most frequently occurring words after removing stop words and common english words as keywords. Dataset details are shown in Table \ref{tdata}.

	\subsection{Keyword based step wise increment}
	\label{keywords}
	First, we experimented using a pointer-generator with a keyword-based log-probability increment at each step. Depending on the category of the document, a list of keywords was determined which needs to be given higher weightage during summarization as mentioned above. In this model, we increase the log-probabilities of these words at each step in the Beam Search decoder according to Eq. \eqref{eqexp} where $T_{ij}$ is the log-probabilities generated by the decoder at step $i$ and for $j^{th}$ word token in the vocabulary. After thorough experimentation, we determine the optimal value of $\alpha$ to be 0.1. But since this strategy is biased towards maximum keyword occurrence irrespective of semantic correctness, this leads to incorrect sentence formations (in the 2nd example in Table \ref{t1} the word `lakhs' is generated twice).
	\begin{minipage}{0.95\linewidth}
	\begin{center}
		\begin{equation}
		P_{ij}=T_{ij} * \alpha \label{eqexp}
		\end{equation}
	\end{center}
\end{minipage}
	
	\subsection{Keyword based end hypothesis increment}Next, we increase the log-probabilities of keywords only at the final step of Beam Search for all the end hypotheses instead of at each step according to the \eqref{eqexp}. This gives better results than the previous strategy, but errors persist (in 1st example, ``passbook contribution" is predicted instead of ``EPF contribution" which is wrong). To solve these issues, we propose a novel strategy of using an Language Model (LM) with Beam Search.

	\section{Proposed Adaptive Beam Search}
	\label{ABS}
	
	\begin{figure}[ht]
		\vskip 0.2in
		\begin{center}
			\centerline{\includegraphics[width=\linewidth]{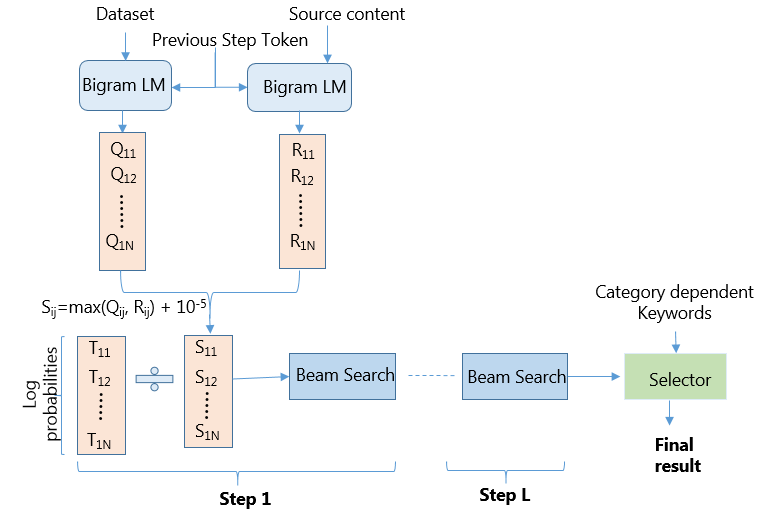}}
			\caption{Adaptive Beam Search for optimized summarization}
			\label{beam-search2}
		\end{center}
		\vskip -0.2in
	\end{figure}
		\begin{figure*}[b]
	\vskip 0.2in
	\begin{center}
		\centerline{\includegraphics[width=0.8\textwidth]{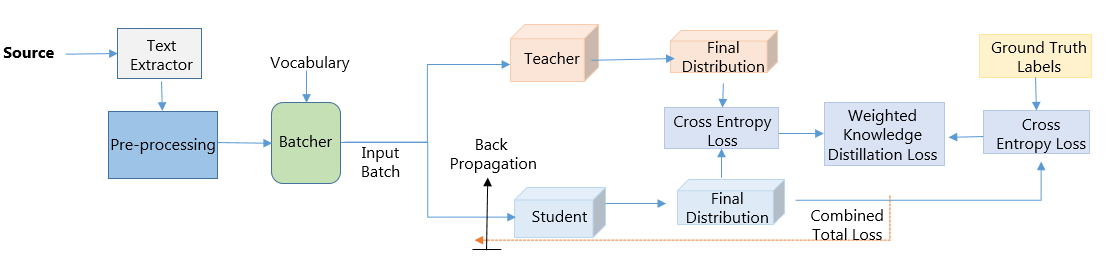}}
		\caption{Pipeline used for training the final summarization model}
		\label{pipeline}
	\end{center}
	\vskip -0.23in
\end{figure*}

\begin{table}[t]
	\caption{In-house Dataset Details}
	\label{tdata}
	\vskip 0.15in
	\begin{center}
		\begin{tabular}{m{0.2\columnwidth}|m{0.6\columnwidth}}
			\toprule
			\textbf{Source} &\textbf{ Categories} \\
			\midrule
			SMS&  Travel tickets, Bank transactions, verification code, advertisements, miscellaneous 
			\\
			\hline
			\vskip 0.05in
			Documents & \vskip 0.05in Travel, Personal information, Receipts, papers/books , Miscellaneous	\\
			
			\bottomrule
		\end{tabular}
	\end{center}
	\vskip -0.1in
\end{table}

	To solve the issue of inconsistent summaries generated through the above search strategy, we use a bigram language model to improve the quality of the final summary generated and also propose a new scoring strategy to select the best hypothesis at the end.
	
	%	\begin{minipage}{.8\linewidth}
		\begin{algorithm}[h]
			\caption{ Final Summary Selection}
			\label{alg}
			\begin{algorithmic}
				\STATE {\bfseries Input:} List of sorted hypotheses $H$\\ at the final step of Beam Search,\\ List of keywords $L$
				\STATE Initialize $F = 0 , S= \emptyset$.
				\FOR{$h_i$ {\bfseries in} $H$}
				\STATE Initialize $f_i = 0 $.
				\FOR{$w_j$ {\bfseries in} $h_i$}
				\IF{$w_j \in L$}
				\STATE $f_i++$
				\ENDIF
				\ENDFOR
				\STATE $S.append (f_i)$
				\IF{$f_i > F$}
				\STATE $F = f_i$
				\ENDIF
				\ENDFOR
				\FOR{$h_i${\bfseries in} $H$}
				\IF{$f_i = F$ {\bfseries or} $f_i = (F - 1)$}
				\STATE{\bfseries return} ($h_i$)
				\ENDIF
				\ENDFOR
			\end{algorithmic}
		\end{algorithm}
	%\end{minipage}

	We enhance the Beam Search by connecting a bigram language model shown in Fig. \ref{beam-search2}. The dataset given as input to the language model is different for each category of text source. The model calculates the probability of bigram occurrence which is then added to the log probabilities from the decoder as per Eq. \eqref{eq1}
	\begin{minipage}{0.95\linewidth}
	\begin{center}
		\begin{equation}
		P_{ij}=\frac{T_{ij}} { \left (max\left (Q_{ij}, R_{ij}\right)+ 10^{-5}\right)}\label{eq1}
		\end{equation}
	\end{center}
\end{minipage}
\vskip 0.08in

	where $Q_{ij}$ is the bigram probability of token predicted in the $(i-1)^{th}$ step and the $j^{th}$ token from vocabulary using input dataset. In case $i$ is 1, then the `$start$' token is considered as the previous step token. Similarly, $R_{ij}$ is calculated using bigrams taken from the source content and $T_{ij}$ is the log-probabilities generated by the decoder at step $i$ and for $j^{th}$ word token in the vocabulary. We use the source bigrams to consider cases like bank statements where the amount is unique in each transaction. Thus, this captures the grammatical correctness of bigrams in source content even if it is not seen previously in the dataset.

	We add a small quantity of $10^{-5}$ to avoid division by 0.  The Selector uses the method described in Algorithm \ref{alg} to select the final summary. $L$ is the list of common and category-specific keywords (explained in Section \ref{keywords}) for the given source and $H$ is the list of all hypotheses ($h_i$ where $0<i\leq beam\ width$) sorted according to the average log-probabilities in the last step of Beam Search. Each hypothesis $h_i$ contains $j$ word tokens represented by $w_j$. $F$ is the global maximum frequency of keywords (in $L$) considering all hypotheses in $H$ i.e,\ $F= max (f_i) \forall f_i \in S$ where $S$ is the set of all $f_i$, the individual keyword frequency of $h_i$. We select the first hypothesis with $f_i =F$ or $f_i = (F-1)$ as the final summary. We have included (F-1) hypotheses also to allow other summaries with a significant number of keywords with higher log-probability to get chosen. The length of the final summary is dependent on the length of the source text content. As it is found that users would make the majority of decisions using the longer levels of summary (15\% and 30\%) according to \cite{sweeney2006effective}, after experimenting, we found 15\% to be optimal. Thus, we use the minimum length to be 15\% of the source length and the maximum length to be 35\% of the source length. Also, the length of the summary is one of the parameters that can be configured by the user (To enable flexible summary size).
	
	\section{Overall Pipeline}	
	From low complexity simple rule-based extractive models to heavily parameterized Transformer based abstractive models, there are various approaches for summarizing text. To choose the most optimal approach for on-device summarization is a challenge as the requirements and environment during deployment can be significantly different from those during the development phase as stated in \cite{lai2018rethinking}. There exists a trade-off between model performance and model size. The pointer-generator network showed promising results on summarization and deploying this architecture on-device is plausible as the model size is relatively small, due to which we experiment on architectures using this as the base. The detailed architecture is explained in the following section.

	\subsection{Dataset}
	\label{dataset}
	The model is first trained on CNN-DM dataset \cite{hermann2015teaching} \cite{nallapati2016abstractive} which contains online news articles. The pretrained model is then trained on How2 Dataset \cite{sanabria2018how2} for speech summarization. 
	
	\subsection{Preprocessing}
	We use simple text pre-processing where the words are first converted to lower case and the special characters are separated. The words are then tokenized and indexed according to the vocabulary (which is of size 50K, determined to optimize final model size). With a batch size of 16, each article is padded to ensure that the input is of the same length and then fed to the model.

	\subsection{Adaptive Beam Search (ABS)}
	We use our proposed ABS strategy along with pointer-generator network \cite{see2017get} to decode during inferencing to adapt to the type of data source. We make use of source-specific LM to enable the model to adapt to the source the details of which are explained in Section \ref{final author}. 
	
\subsection{Knowledge Distillation}
\label{author info}
Knowledge distillation(KD) \cite{hinton2015distilling} is a technique to compress large models.
We experiment to perform KD on our model by using a temperature of $1$. Considering $\alpha$ to be equal to $1-\beta$ the final loss function for the student model is described in \eqref{eq4}.
	\begin{minipage}{0.95\linewidth}
\begin{center}
	\begin{equation}
	Loss=\left(1 - \beta \right)*{Loss_{target}} + \beta * Loss_{t-s} \label{eq4}
	\end{equation}
\end{center}
\end{minipage}
\vskip 0.1in
$Loss_{t-s}$ is the cross-entropy loss of the student and teacher distributions and $Loss_{target}$ is the cross-entropy loss considering target summary.
 We consider the value of $\beta$ to be $0.4$ as our intention was for the model to be slightly more biased towards the ground truth. The teacher architecture consists of 256 BiLSTM cells, and embedding dimension as 128 with a total of 400 steps. We use a decoder consisting of 100 steps for training. In student architecture, we reduce the number of BiLSTM cells to 150. We train the student directly, using a teacher model trained with coverage loss which excludes the need to train the student with coverage loss again. The final training pipeline is shown in Figure \ref{pipeline}.

\subsection{On-device deployment}
For deploying on-device, we first performed quantization using TensorFlow tflite as well as graph transform tool on the student model to further reduce the size, but we observed a significant deterioration in the performance. Thus, we do not use the quantized model. We implement preprocessing and beam search decoding on Android as the model graph after freezing does not support these operations. We build the TensorFlow \cite{tf}(version 1.10) library from source including some operations previously not included(like ScatterND) in the default shared object file and use the built shared object file to facilitate model inference on-device. 

\begin{table*}[ht]
	\caption{SMS summarization results benchmarked with BERT results highlighting key tokens (in \textcolor{blue}{blue}) necessary in summary}	\label{t2}
	\begin{center}
		\begin{small}
			
			\begin{tabular}{m{0.3\textwidth}|m{0.2\textwidth}|m{0.2\textwidth}|m{0.2\textwidth}}
				\hline
				\vskip 0.025in
				\textbf{Source} & 	\vskip 0.025in \textbf{Pointer-Generator with ABS} & 	\vskip 0.025in \textbf{Pointer-Generator with ABS after KD} & 	\vskip 0.025in \textbf{BERT} \\
				\hline
				\vskip 0.025in
				Dear Mr Kumar -Your IndiGo \textcolor{blue}{PNR is AFBQ9P- 6E  369} 04Jan BLR (T1)-HYD, 1435-1550 hrs. Web check-in now by clicking here - http://I9f.in/lf21xGj48j You can print boarding pass from Boarding Pass Printing Units outside Gate 1
				\vskip -0.025in
				&your indigo \textcolor{blue}{pnr is afbq9p} - hyd , 1435-1550 hrs . you can print boarding pass from
				&  mr kumar - your indigo \textcolor{blue}{pnr is afbq9p} - 6e 369 04jan blr 
				& you can print boarding pass from boarding pass printing units outside gate 1
				\\
				\hline
				\vskip 0.025in
				Txn of \textcolor{blue}{INR 209.00} done on Acct XX013 on 15-DEC-19.Info: VPS*KAILASH P.Avbl Bal:INR 50,698.72.Call 18002662 for dispute or SMS BLOCK 013 to 9215676766
				\vskip -0.025in
				& txn of \textcolor{blue}{inr 209.00} done on acct xx013 on 15-dec-19 . info  .  
				&  txn of \textcolor{blue}{inr 209.00} done on acct xx013 on 15-dec-19 . Info 
				& \textcolor{blue}{inr 209.00} done on acct xx013 on 15-dec-1
				
				\\
				\hline
			\end{tabular}
			
		\end{small}
	\end{center}
	\vskip -0.25in
\end{table*}

\begin{table}[ht]
	\caption{Comparative On-device Performance analysis}
	\label{tcomp}
	\begin{center}
		\begin{tabular}{m{0.2\columnwidth}|m{0.2\columnwidth}|m{0.2\columnwidth}|m{0.2\columnwidth}}
			\toprule
			\textbf{Model} & \textbf{Keywords Recalled} & \textbf{Model Size} & \textbf{On-device Inference Time (seconds/character)} \\
			\midrule
			\vskip 0.05in
			Pointer generator & 49\% & 80.07 MB &  0.035 s/char \\
			\hline
			\vskip 0.05in
			BERT & \vskip 0.05in 56\% & \vskip 0.05in $>$2 GB & --$^{\ast}$\\
			\hline
			
			\vskip 0.05in
			Pointer generator with ABS after KD & 69\%  &55.3 MB & 0.03 s/char  \\
			
			\bottomrule
			
			\multicolumn{4}{l}{$^{\ast}$ On-device inference time is not applicable to BERT}
		\end{tabular}
	\end{center}
	\vskip -0.18in
\end{table}

\begin{table}[ht]
	\caption{Average ROUGE F-scores on CNN-DM dataset.}
	\label{t3}
	\begin{center}
		
		\begin{tabular}{m{0.2\columnwidth}|m{0.2\columnwidth}|m{0.2\columnwidth}|m{0.2\columnwidth}}
			\toprule
			\textbf{Model} & \textbf{ROUGE-L} & \textbf{ROUGE-1} & \textbf{ROUGE-2} \\
			\midrule
			Teacher&  30.12$\pm$ 0.4
			&  33.48$\pm$ 0.4&  13.41$\pm$ 0.4 \\
			\hline
			Student before KD &28.11$\pm$ 0.4	& 31.58$\pm$0.4  &  11.43$\pm$ 0.3	\\
			\hline
			Student after KD & 30.18$\pm$ 0.5
			&  33.55$\pm$ 0.5					 
			& 13.36$\pm$ 0.4\\
			
			\bottomrule
		\end{tabular}
		
	\end{center}
	\vskip -0.25in
\end{table}

\begin{table}[ht]
	\caption{Average ROUGE score evaluation on How2 Dataset.}
	\label{t4}
	\begin{center}
		\begin{tabular}{m{0.2\columnwidth}|m{0.2\columnwidth}|m{0.2\columnwidth}|m{0.2\columnwidth}}
			\toprule
			\textbf{Metric} & \textbf{ROUGE-L} & \textbf{ROUGE-1} & \textbf{ROUGE-2} \\
			\midrule
			F-score & 29.95$\pm$ 0.4 & 36.54$\pm$ 0.4 &  15.13$\pm$ 0.4\\
			Recall & 33.46$\pm$ 0.5 & 41.29$\pm$ 0.5 & 17.26$\pm$ 0.3\\
			Precision & 27.31$\pm$ 0.4 & 34.27$\pm$ 0.4 & 14.12$\pm$ 0.4 \\
			
			\bottomrule
		\end{tabular}
	\end{center}
	\vskip -0.25in
\end{table}

\section{Summarization on different sources}
\label{final author}
This model architecture supports the summarization of SMS,  voice messages  and can be extended for documents and URLs as well. For SMS summarization, the LM is trained on SMS data categorized into travel tickets, bank transactions, OTP, advertisements, miscellaneous. Important details that need to be extracted depend upon the category of the SMS. We list out keywords like `dollars', `rupees', `amount' to be common keywords across all categories and convey important details of the message in most cases. Other category-specific keywords like ‘PNR’, ‘Delayed’, etc. are only considered when the SMS belongs to a specific category (in this case `travel') which is determined by another list of keywords. To select the final summary we use Algorithm \ref{alg}.

	For speech summarization, we convert speech to text using ASR, followed by sentence boundary detection, the output is sent to the model which is also trained on the How2 dataset.

	\section{Results and Discussion}

	We use the Samsung Galaxy A50 smartphone model (Android 9.0, 6GB RAM, 64GB ROM, Samsung Exynos 7 Octa 9610) to conduct our experiments.
	
	To evaluate on SMS summarization, we benchmark the results with that of BERT (12-layers, 768-hidden, 12-heads, 110M parameters). We create an in-house dataset explained in \ref{keywords} and use this to calculate the accuracy of keywords extracted in the summaries by each model. We evaluate by calculating the recall of keywords in the final summary. We observe that our model can extract an average of 69\% keywords as compared to 56\% accuracy of BERT and 49\% of pointer-generator without ABS. Some examples are as shown in Table \ref{t2}. The overall performance can be seen in Table \ref{tcomp} which shows our model performs better than BERT and baseline model. The inclusion of an LM trained on the same structured text and inclusion of category-specific keywords improves the performance.
	
			\begin{figure}[ht]
		\begin{center}
			\centerline{\includegraphics[width=0.7\columnwidth]{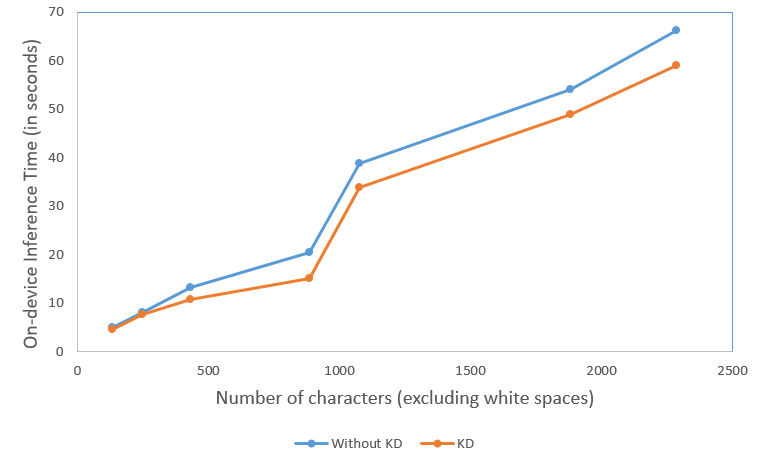}}
			\caption{Graph between inference time and character count}
			\label{graph}
		\end{center}
		\vskip -0.2in
	\end{figure}

	To evaluate the impact of Knowledge Distillation (KD), we calculate the  ROUGE scores on 4000 randomly sampled test articles of CNN-DM dataset on the teacher, student model trained without and with KD, the F-scores of which are as shown in Table \ref{t3}. The results prove that there is a strong improvement in the model performance after KD. We observe a reduction in model size by 30.9\% after KD.
	
	Fig. \ref{graph} shows the on-device inference time as a function of the character count for the model trained without and with KD. The latter has better inference time (reduction in model loading time:60.5\%, average inference time:12.66\%).

	To further check the efficacy of the student model after KD on speech, we also evaluated the performance of our model on the How2 dataset as shown in Table \ref{t4}.
	
	\begin{figure}[ht]
		\begin{center}
			\centerline{\includegraphics[width=0.75\columnwidth]{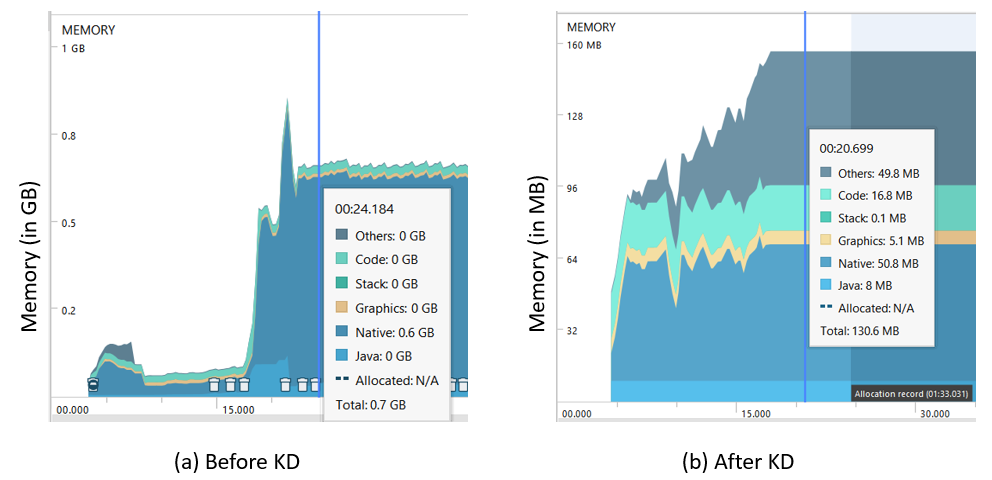}}
			\caption{Comparision between RAM used before and after KD}
			\label{ram}
		\end{center}
	\vskip -0.25in
	\end{figure}

	There is a significant reduction in on-device RAM usage as well which is measured using Android Memory Profiler. The model utilizes a total RAM of 700MB before KD which is reduced to 130.6MB after KD. Fig. \ref{ram} captures the RAM usage for two scenarios: with and without KD. The memory segments are captured at a particular instant of time while on-device summarization is in progress for the same sample. 
	
	The above results prove that our model adapts to short text, long text, and speech with the help of ABS.

	\section{Conclusion and Future work}
	In this paper, we propose a novel Adaptive Beam Search strategy and a new scoring technique for optimal on-device summarization. The proposed model can adapt to different types of text and speech sources on-device which makes it highly useful to mobile users over the globe. We show that our model extracts more keywords as compared to BERT by 13\% with a 97.6\% decrease in model size. We also show that Knowledge Distillation helps reduce the model size by 30.9\%. In the future, it will be intriguing to explore other model compression techniques and test their efficiency for this task.  
	\bibliographystyle{IEEEtran}
	
	\bibliography{IEEEexample}

\end{document}